\documentclass[preprint,amsmath,amssymb]{revtex4}

\usepackage{amsmath,amssymb,amsthm}
\usepackage{graphicx}
\usepackage{epsfig}

\begin{document}

\title{No one-hidden-layer neural network can represent multivariable functions}

\author{M. Inoue$^1$, M. Futamura$^2$, \ and H. Ninomiya$^{1,2}$}
\affiliation{$^1$ School of Interdisciplinary Mathematical Sciences, Meiji University, \\
4-21-1 Nakano, Nakano-ku, Tokyo 164-8525, Japan\\
$^2$ Graduate School of Advanced Mathematical Sciences, Meiji University, \\
4-21-1 Nakano, Nakano-ku, Tokyo 164-8525, Japan}

\date{\today}

\begin{abstract} 
In a function approximation with a neural network, an input dataset is mapped to an output index by optimizing the parameters of each hidden-layer unit. 
For a unary function, we present constraints on the parameters and its second derivative by constructing a continuum version of a one-hidden-layer neural network with the rectified linear unit (ReLU) activation function. 
The network is accurately implemented because the constraints decrease the degrees of freedom of the parameters. 
We also explain the existence of a smooth binary function that cannot be precisely represented by any such neural network. 
\end{abstract}

\maketitle

\section*{Introduction}

Machine learning using multilayer artificial neural networks has made rapid progress over the past decades, and it has been successfully applied in various fields~\cite{LBH, SJ, SS, BBU, ZEK, SB, OL, LS18, MBW}. 
One of its major and long-standing applications is function approximation~\cite{Cy, F, LS}.  
Learning algorithms can be considered to provide a function that maps an input dataset to an output index. 
Neural networks provide function approximations when the functions are not given {\it a priori}.
It has been shown that a given continuous function on a compact set can be approximately realized by a one-hidden-layer feedforward neural network~\cite{Cy, F, HSW1, HSW2, P}. 

One of the main concerns in function approximation is to estimate the number of neurons (i.e., the units in hidden layers).
 It is widely believed that a neural network with a larger number of hidden layers and their units yields more precise approximations~\cite{LS}. 
Eldan and Shamir, for example, showed that, to approximate a function, a one-hidden-layer network requires an exponential number of neurons, whereas a two-hidden-layer network requires a polynomial number of neurons~\cite{ES}. 
Most of these studies, however, show the existence of approximating networks using nonconstructive methods. 
Many optimization methods have been proposed, but it is unclear what the chosen parameter values mean. 
As a neural network becomes larger and deeper, the operations of its hidden-layer units become more complicated. 

To understand the learning mechanisms systematically and what the black box of the hidden layers implements, it is necessary to determine how a (neural) network that can approximate a given function be constructed. Then, we can understand what the network optimizes during the training process. 
Suzuki proposed constructive approximations using feedforward neural networks; however, the proposed method is complicated~\cite{S}. 

\section*{Materials and methods}

\begin{figure}[h]
\includegraphics[width=100mm]{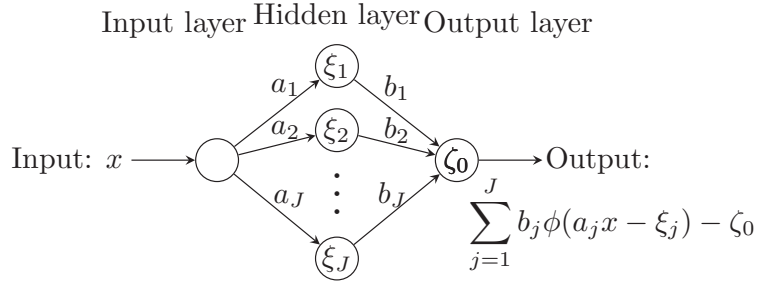}
\caption{{\bf Schematic view of one-hidden-layer neural network model.}
}
\label{fig:network}
\end{figure}

Integral representation is a powerful tool to formulate the aforementioned problem~\cite{M}. 
The feedforward one-hidden-layer neural network with an activation function $\phi$ is defined by 
\begin{equation}
\label{eq:nn}
f(x)=\sum_{j=1}^Jb_j\phi(a_j x-\xi_j) - \zeta_0
\end{equation}
where $a_j$/$b_j$ denotes the connection weights between input/output units and the $j$th unit in the hidden layer; $\xi_j$ denotes the bias of the $j$th unit $(j=1,\cdots, J)$; $\zeta_0$ denotes the bias of the output unit (Fig~\ref{fig:network}). 
Considering the continuum extension of the discrete neural network, Murata showed that such a neural network can be obtained by discretizing its integral representation
\begin{eqnarray*}
h(x)=\int_{D} b(a,\theta)\phi(a x+\theta)d\mu(a,\theta)
\end{eqnarray*}
where $b$ and $\mu$ denote a continuous function and a measure on $D\subset \mathbb{R}^2$, respectively~\cite{M}. 
This integral representation can be regarded as the dual ridgelet transformation~\cite{SM,SM2}. 

Motivated by these works, we introduce a natural integral representation of Eq~(\ref{eq:nn}) over a one-dimensional interval instead of a two-dimensional set $D$. 
Throughout this letter, we adopt the ReLU function as the activation function (i.e., $\phi(x) \equiv \max\{x,0\}$). 
We propose the constructive manner of function approximations by using the continuum version of neural networks. 
In this study, we modified the integral representation presented in~\cite{M,SM}. 
We emphasize that domain $D$ of the feedforward neural network is one-dimensional (i.e., each parameter set $(a_j,b_j,\xi_j)$ is a function of $\xi_j$). 
We also show the existence of a smooth function with two variables that cannot be approximated by any continuum one-hidden-layer neural network. 

For any smooth function $f$, the integration by parts and the Fubini theorem yield 
\begin{eqnarray*}
 f(x)&=&f(0)+\int_0^xf'(y)dy\\
 &=&f(0)+f'(0)x + \int_{0}^{x} \int_{0}^{y}f''(\xi)d\xi dy\\
 &=& f(0)+f'(0)x + \int_{0}^{x} f''(\xi)(x-\xi)d\xi.
\end{eqnarray*}
This simple calculation implies that by setting $\zeta_0=-f(0)$, $b_1=f'(0)$, and $g=f''$, we obtain 
\begin{equation}
\label{eq:f-int}
 f(x)=-\zeta_0+b_1\phi(x)+ \int_{0}^{L}g(\xi)\phi(x-\xi)d\xi.
\end{equation}
In fact, substituting $x=0$ in Eq~(\ref{eq:f-int}) yields $f(0)=-\zeta_0$. 
Differentiating Eq~(\ref{eq:f-int}) with respect to $x$, we obtain $f'(x)=b_1+ \int_{0}^{x}g(\xi)d\xi$. 
Similarly, we get $b_1=f'(0)$ and $g(x)=f''(x)$ for $0 \leq x \leq L$. Thus, the integral representation, Eq~(\ref{eq:f-int}), is uniquely determined for any smooth function $f$. 

Next, we consider the correspondence between this representation and the neural network. 
Let $\Delta_{J,\xi} \equiv \{\xi_j\}_{j=1}^J$ be a division of the interval $[0,L]$ consisting of points $0=\xi_0<\xi_1<\cdots<\xi_J= L$. 
We introduce the following one-hidden-layer neural network: 
\begin{eqnarray}
F(x;\Delta_{J,\xi})& \equiv&
f(0)+f'(0)\phi(x) \nonumber\\
&& +\sum_{j=0}^{J-1}f''(\xi_j)(\xi_{j+1}-\xi_{j})\phi(x-\xi_j).
 \label{def:F}
 \end{eqnarray}
Subtracting Eq~(\ref{eq:f-int}) and Eq~(\ref{def:F}) yields
\begin{eqnarray*}
\lefteqn{|f(x)-F(x;\Delta_{J,\xi})| }\\
 &=& | \sum_{j=0}^{J-1} \int_{\xi_j}^{\xi_{j+1}} {\biggl(} f''(\xi)\phi(x-\xi)-f''(\xi_j)\phi(x-\xi_j) {\biggr)} d\xi | \\
&\leq&\sum_{j=0}^{J-1} \int_{\xi_j}^{\xi_{j+1}}|f''(\xi)-f''(\xi_j)|\phi(x-\xi)d\xi\\
 &&+\sum_{j=0}^{J-1} |f''(\xi_j)|\cdot{\Biggl|}\int_{\xi_j}^{\xi_{j+1}}{\biggl(}\phi(x-\xi)-\phi(x-\xi_j) {\biggr)} d\xi{\Biggr|}\\
&\leq&\sum_{j=0}^{J-1} \|f'''\|_{C_{0}} |\xi_{j+1} - \xi_{j}| L(\xi_{j+1}-\xi_j)\\
&& + \|f''\|_{C^0}|\Delta_{J,\xi}| \sum_{j=0}^{J-1} \dfrac 12(\xi_{j+1}-\xi_j)\\
&\leq&\Big(L^2\|f'''\|_{C^{0}}+\dfrac L2\|f''\|_{C^0}\Big) |\Delta_{J,\xi}|
 \end{eqnarray*}
with $|\Delta_{J,\xi}|\equiv \max_{0\leq j \leq J-1}|\xi_{j+1}-\xi_{j}|$ and $\|h\|_{C^0}\equiv \max_{0\leq x \leq L}|h(x)|$ for any continuous function $h$ over $[0,L]$. 

\begin{figure}[h]
\includegraphics[width=100mm]{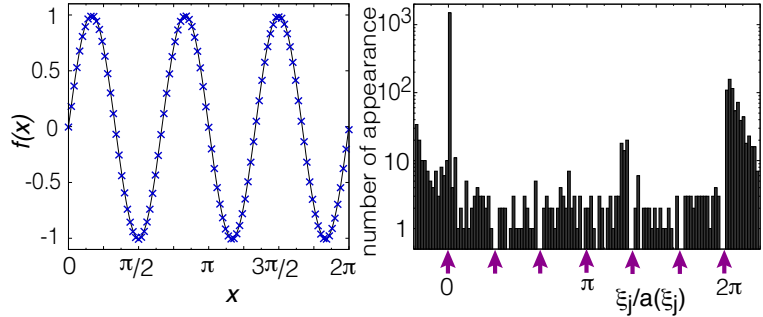} 
\caption{{\bf Approximation of the sine function and distributions of $\xi_j / a(\xi_j)$.}
$f(x)=\sin{3x} \ (0 \leq x \leq 2\pi)$ (left, solid line) is trained by a one-hidden-layer neural network with $J=3,000$ units and $50,000$ training data. 
The trained result is over written with $\times$ for $x=2\pi j/100$ $( j =0, 1, \cdots 100)$.
The maximum error is $0.0274$. 
The distribution of $\xi_j / a(\xi_j)$ (right) is sparse at $\xi_j \sim k \pi/3 \ (k=0,1,\cdots, 6)$ because $f''(\xi_j)=-3^2\sin{3\xi_j}$ is sufficiently small, and the estimated approximation error is negligible even with a larger $(\xi_{j+1} - \xi_j)$. 
}
\label{fig:2}
\end{figure}

Summarizing the above, we note that for any smooth function $f(x)$ defined on $0 \leq x \leq L$ and a division $\Delta_{J,\xi}$, there is a positive constant $C_1$ depending only on $L$ and $f$ such that, for $0\le x\le L$, 
 \begin{eqnarray}
 \max_{0\leq x \leq L} |f(x)-F(x;\Delta_{J,\xi})|\leq C_1|\Delta_{J,\xi}|.
\label{ineq:error-F}
 \end{eqnarray}
Thus, $F(x;\Delta_{J,\xi})$ is realized by a neural network given by Eq~(\ref{eq:nn}) with $a_j=1,\ b_j=f''(\xi_j)(\xi_{j+1}-\xi_{j})$, and $\xi_j=Lj/J$. 
From the above argument, $f(x)$ can be approximated by Eq~(\ref{def:F}). However, the influences of the terms of Eq~(\ref{def:F}) depend on the coefficients $f''(\xi_j)(\xi_{j+1}-\xi_{j})$. 
Namely, when $b_j$ is small, the term $b_j \phi(a_j x-\xi_j)$ is negligible, which is observed in Fig~\ref{fig:2}.
To study the deeper relation between the representation in Eq~(\ref{def:F}) and the neural network in Eq~(\ref{eq:nn}), we consider the coefficients of the terms in Eq~(\ref{def:F}) next. 
The significant difference between them is that Eq~(\ref{def:F}) does not include the case where $a_j<0$. 

By considering that the activation function $\phi$ is a ReLU function, Eq~(\ref{eq:nn}) is rewritten as 
\begin{eqnarray*}
\sum_{j=0}^Jb_j|a_j|\phi\left((-1)^{\ell_j}\Big(x-(-1)^{\ell_j}\dfrac {\xi_j}{|a_j|}\Big)\right) -\zeta_0
\end{eqnarray*}
where $\ell_j$ is $0$ or $1$. 
Therefore, the one-hidden-layer neural network with the activation function $\phi$ is represented as 
\begin{equation}
\label{eq:nn2}
b_0+\sum_{j=1}^J\Big[b^+_j\phi(x-\xi_j)+b^-_j\phi(\xi_j-x)\Big] -\zeta_0.
\end{equation}

To include the case where $a_j<0$, 
we extend Eq~(\ref{eq:f-int}) to 
\begin{eqnarray}
\label{eq:f-int2}
 f(x)&\!\!=\!\!&-\zeta_0+b_1^+\phi(x)+b_1^-\phi(L-x)\nonumber\\
&& + \int_{0}^{L}\Big[g^+(\xi)\phi(x-\xi)+g^-(\xi)\phi(-x+\xi)\Big]d\xi.\qquad\label{eq:bnn}
\end{eqnarray}
Here, it is called a ``(one-hidden-layer) continuum neural network''. 
We note that this includes Eq~(\ref{eq:f-int}) as a special case with $b_1^-=0$ and $g^- = 0$. 
Differentiating Eq~(\ref{eq:f-int2}) twice with respect to $x$, we also obtain $f''(x)=b^+(x)+b^-(x)$ for $0<x<L$. 

\begin{figure}[h]
\includegraphics[width=100mm]{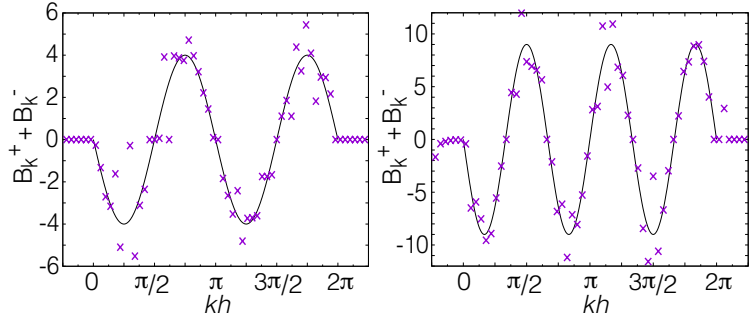}
\caption{{\bf Relation between the bias and the connection weights of hidden-layer units.}
$kh$ (abscissa) and $B^+_k+B^-_k$ (ordinate) are plotted for $f(x) = \sin{2x}$ (left) and $f(x) = \sin{3x}$ (right) with $h=2\pi/50$ and $L=2\pi$. 
$f(x) = \sin{2x}$ is trained by a one-hidden-layer neural network with $J=3,000$ units and $30,000$ training data, and the same data as those presented in Fig~\ref{fig:2} are used for $f(x) = \sin{3x}$. 
Please refer to the text for the axis-label definitions. The solid lines indicate the theoretical results in Eq~(\ref{eq:tmp}), i.e., $f''(x)=-M^2\sin Mx$. 
}
\label{fig:plotB}
\end{figure}

A similar argument guarantees the function approximation of Eq~(\ref{eq:nn2}) corresponding to Eq~(\ref{def:F}). 
Here, we confirm the relation between Eq~(\ref{eq:nn2}) and Eq~(\ref{eq:bnn}) numerically. 
For a given $f(x)$, suppose that we have $(a_j,b_j,\xi_j, \zeta_0)$ for $j=1,\cdots,J$ in Eq~(\ref{eq:nn}). 
Then, Eq~(\ref{eq:nn}) becomes
\begin{eqnarray}
\lefteqn{b_0+\sum_{j=1}^Jb_j\phi(a_j x-\xi_j) -\zeta_0}\nonumber\\
&=&
-\zeta_0+\sum_{k=1}^K\sum_{kh<\xi_i/a_i<(k+1)h}b_i|a_i|\phi\left(\dfrac {a_i}{|a_i|}\Big( x-\dfrac {\xi_i}{a_i}\Big)\right)\nonumber\\
&=&
-\zeta_0+\sum_{k=1}^K\sum_{kh<\xi_i/a_i<(k+1)h,\ a_i\ge 0}b_i|a_i|\phi\left( x-\dfrac {\xi_i}{a_i}\right)\nonumber\\
& &+\sum_{k=1}^K\sum_{kh<\xi_i/a_i<(k+1)h,\ a_i< 0}b_i|a_i|\phi\left( \dfrac {\xi_i}{a_i}-x\right).\label{eq:BNN0}
\end{eqnarray}
For $0 \leq x \leq L$, $\phi\left( x- \xi_i / a_i \right)=0$ and $\phi\left( \xi_i / a_i -x\right)= \xi_i / a_i -L+\phi(L-x)$ when $\xi_i/a_i>L$, and $\phi\left( x- \xi_i / a_i \right)=- \xi_i / a_i +\phi(x)$ and $\phi\left(\xi_i / a_i -x\right)=0$ when $\xi_i/a_i<0$. Therefore, we assume that $0\le \xi_i/a_i\le L$. 
For a small positive constant $h$, we set $B_k^+ \equiv \sum_{kh<\xi_i/a_i<(k+1)h,\ a_i\ge 0}b_i|a_i|/h$ and $B_k^+ \equiv \sum_{kh<\xi_i/a_i<(k+1)h\ a_i<0}b_i|a_i|/h$ for $k=0,1, \cdots, K=L/h$.  
Then, we  obtain 
\begin{eqnarray}
\lefteqn{-\zeta_0+\sum_{j=1}^Jb_j\phi(a_j x-\xi_j)}\nonumber\\
&\approx &-\zeta_0+\sum_{k=1}^K B_k^+h\phi\left( x-kh\right)
+\sum_{k=1}^K B_k^-h\phi\left(kh-x\right)\qquad \label{eq:BNN}
\end{eqnarray}
where $\xi_i/a_i\approx kh$ is used. 
By letting $h\to 0$, Eq~(\ref{eq:BNN}) converges to Eq~(\ref{eq:bnn}) with $B_k^\pm\approx b^\pm (kL/K)$, i.e., 
\begin{eqnarray}
B_k^++B_k^-\approx b^+ (kL/K)+b^- (kL/K)=f''(kL/K) \label{eq:tmp}
\end{eqnarray}
as shown in Fig~\ref{fig:plotB}. 

\section*{Results and Discussion}

Next, we consider the neural network approximating a function with $m (\geq 2)$ variables. 
The natural extension of the (one-hidden-layer) continuum neural network to a multivariable case is 
\begin{equation}
\label{eq:m-abnn}
 f({\bf x})=-\zeta_0+\sum_{j=1}^Jb_{1,j}\phi({\bf c}_j{\bf x})+ \int_{0}^{L}b(\xi)\phi({\bf a}(\xi){\bf x}-\xi)d\xi, 
\end{equation}
where ${\bf x}=(x_1,\cdots,x_m)$, ${\bf c}_j=(c_{1,j},\cdots,c_{m,j})$, and ${\bf a}(\xi)=(a_1(\xi),\cdots,a_m(\xi))$. 

Now, we explain that there is a smooth function $f$ that cannot be realized by any one-hidden-layer continuum neural network, considering $f(x,y) = xy$ as an example. 
Here, we outline the proof. 
Suppose that $(b_1,c_1,c_2, L,a_1,a_2,b, \zeta_0)$ satisfies Eq~(\ref{eq:m-abnn}) for any $(x,y)$ in $0 \leq x \leq L, 0 \leq y \leq L$. 
We divide the situation into the following four cases and arrive at a contradiction: (i) $a_1(0)< 0,\ a_2(0)< 0$, (ii) $a_1(0)>0$, (iii) $a_2(0)>0$, and (iv) other cases. 

For case (i), 
there is a point $(x_0,y_0)$ that satisfies $a_1(\xi)x_0+a_2(\xi)y_0-\xi<0$ for $0\le \xi\le L$. 
We can assume that $c_{1,j}x+c_{2,j}y \neq 0$ and $a_1(\xi)x_0+a_2(\xi)y_0-\xi<0$ in some neighbourhood of $(x_0,y_0)$. 
Thus, Eq~(\ref{eq:m-abnn}) becomes 
\begin{equation}
  xy = -\zeta_0 + \sum_{j=1}^Jb_{1,j}\phi(c_{1,j}x + c_{2,j}y)\quad \nonumber
\end{equation}
in that neighbourhood, which is a contradiction. 

Next, we consider case (ii). 
Similar to case (i), we note that $a_1(0)x+a_2(0)y>0$ in the neighbourhood of some point $(x_0,y_0)$. 
Because $\xi-a_1(\xi)x-a_2(\xi)y$ is a monotonically increasing function of $\xi$, 
there is a unique function $\eta=\eta(x,y)$ that satisfies
\begin{equation}
  \eta=a_1(\eta)x+a_2(\eta)y  \nonumber
\end{equation}
for any $(x,y)$ in the neighbourhood. 
By taking a smaller neighbourhood, if necessary, 
we assume that $\eta(x,y)<L$ and that $c_{1,j}x + c_{2,j}y>0$ for $j=1,\cdots, J_0$ and $c_{1,j}x + c_{2,j}y\le 0$ for $j=J_0+1,\cdots,J$. 
Then, Eq~(\ref{eq:m-abnn}) becomes 
\begin{eqnarray*}
 xy&=& -\zeta_0 + \sum_{j=1}^{J_0}b_{1,j}(c_{1,j}x + c_{2,j}y) \\
 &&+ \int_{0}^{\eta(x,y)} b(\xi)(a_{1}(\xi) x + a_{2}(\xi)y - \xi)d\xi. 
\end{eqnarray*}
in that neighbourhood. 
Differentiating the above equation twice with respect to $x$ and $y$  yields 
\begin{eqnarray}
 b(\xi)a_{1}(\xi) \eta_{x} (x,y)&=& 0,\label{eq-ba1}\\
 b(\xi)a_{1}(\xi) \eta_{y}(x,y) &=& 1,\label{eq-ba2}\\
 b(\xi)a_{2}(\xi) \eta_{x}(x,y) &=& 1,\label{eq-ba3}\\
 b(\xi)a_{2}(\xi) \eta_{y}(x,y) &=& 0.\label{eq-ba4}
\end{eqnarray}
Multiplying both sides of Eq~(\ref{eq-ba1}) by Eq~(\ref{eq-ba4}) and Eq~(\ref{eq-ba2}) by Eq~(\ref{eq-ba3}), we arrive at a contradiction. 
Similarly, we can arrive at contradictions in the other cases.

\section*{Conclusions}

In this letter, we introduced continuum neural networks Eq~(\ref{eq:f-int}) and Eq~(\ref{eq:bnn}) using integral representations. 
With the condition of Eq~(\ref{ineq:error-F}), the feedforward neural network (Eq~(\ref{eq:nn})) can be given by the discretized version (Eq~(\ref{def:F})) of the continuum neural network. 
This also shows the relationship between the units' parameters and the approximation function.
Our interpretation of the neural network is simple compared with that proposed in a previous study~\cite{S}. 
According to our constructive analysis, $\xi_j/a_j$ must take a value within the domain of the function, and $a_j \approx 0$ is an irrelevant choice because the corresponding unit can be replaced by a bias. 
Moreover, $b_j |a_j|$ is also restricted by the dependence relationship given by Eq~(\ref{eq:tmp}). 
Thus, each parameter set $(a_j,b_j,\xi_j)$ must be a function of $\xi_j$ as $(a_j (\xi_j), b_j (\xi_j), \xi_j)$, and each $\xi_j/a_j$ must be arranged while balancing with the others. 
More precisely, it is preferable that $\xi_j/a_j$ are uniformly distributed at intervals, especially at the points where $f''\ne 0$. 
With these constraints on the parameters, it is easy to construct a neural network when approximating a known one-variable function. 

\begin{figure}[h]
\includegraphics[width=100mm]{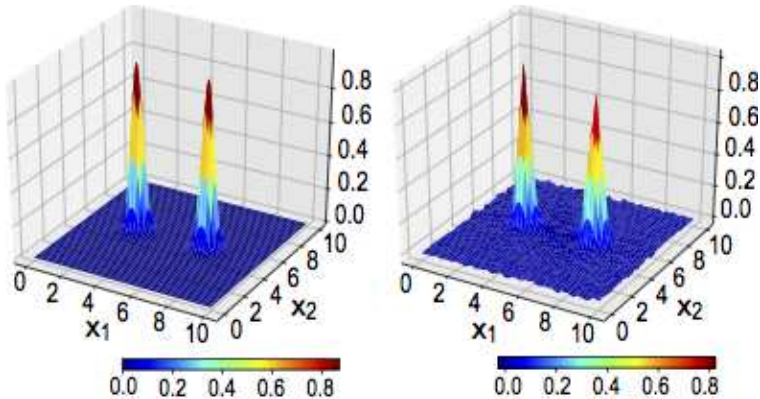}
\caption{{\bf Approximation of a multivariable function by a one-hidden-layer neural network.}
The training data and trained results are presented on the left and right sides, respectively.  
 $f(x,y)=e^{-a(x-x_1)^2-a(y-y_1)^2}+e^{-a(x-x_2)^2-a(y-y_2)^2}$ with $a=5.0, x_1 =3.0, x_2 =7.0, y_1 = y_2 = 5.0$ and $L_x = L_y = 10$ is machine-learned using 1,000 hidden layer units and 250,000 training data. 
 The maximum error is 0.138, and the mean squared error is 0.0102. 
 }
\label{fig:4}
\end{figure}

In contrast, for the approximation of a multivariable function, there are no such constraints. 
We showed the existence of a multivariable function that cannot be represented by any one-hidden-layer ``continuum neural network''. 
However, this result does not imply that such a multivariable function cannot be approximated by any one-hidden-layer neural network. 
In fact, such an approximation is numerically realized with some approximation error (Fig~\ref{fig:4}). 
This is the reason why the approximation of a multivariable function with a one-hidden-layer neural network is difficult and causes an explosion in the number of units. 

If we consider neural networks with independent parameters or those with multiple hidden layers,
the situation will be different. 
For the former case, Sonoda and Murata have given the integral representation even for the approximation of a multivariable function by extending the ridgelet transform to a ReLU function~\cite{SM}. 
For the latter case, for example, Eldan and Shamir have shown the power of the depth of a neural network by comparing the required number of units between one- and two-hidden-layer networks~\cite{ES}. 
We can apply our method to multiple-hidden-layer neural networks, although it is complicated. 
This will be an interesting topic for future work.


\begin{thebibliography}{30}

\bibitem{SS}
Saad D, Solla SA.
\newblock {{E}xact solution for on-line learning in multilayer neural networks}.
\newblock Phys. Rev. Lett. 1995 May 22;74(21):4337--4340. 

\bibitem{BBU} 
Bunzmann C, Biehl M, Urbanczik R.
\newblock {{E}fficiently learning multilayer perceptrons}.
\newblock Phys Rev Lett. 2001 Mar 5;86(10):2166--2169. 

\bibitem{ZEK}
Rosen-Zvi M, Engel A, Kanter I.
\newblock {{M}ultilayer neural networks with extensively many hidden units}.
\newblock Phys Rev Lett. 2001 Aug 13;87(7):078101.

\bibitem{SB}
Sussillo D, Barak O.
\newblock {{O}pening the black box: Low-dimensional dynamics in high-dimensional recurrent neural networks}.
\newblock Neural Comput. 2013 Mar;25(3):626--649

\bibitem{LBH} 
Cun YL, Bengio Y, Hinton G.
\newblock {{D}eep learning}.
\newblock Nature. 2015 May 28;521(7553):436--444

\bibitem{SJ} 
Schmidhuber J.
\newblock {{D}eep learning in neural networks: An overview}.
\newblock Neural Netw. 2015 Jan;61:85--117. 

\bibitem{OL} 
Espinosa-Ortega T, Liew TCH.
\newblock {{P}erceptrons with Hebbian learning based on wave ensembles in spatially patterned potentials}.
\newblock Phys Rev Lett. 2015 Mar 20;114(11):118101.

\bibitem{LS18} 
Li B, Saad D.
\newblock {{E}xploring the function space of deep-learning machines}.
\newblock Phys. Rev. Lett. 2018 120(24):248301. 

\bibitem{MBW} 
Mehta P, Bukov M, Wang C-H, Day AGR, Richardson C, Fisher CK, Schwab DJ.
\newblock {{A} high-bias, low-variance introduction to Machine Learning for physicists}.
\newblock Phys. Rep. 2019 810:1--124.

\bibitem{Cy} 
Cybenko G.
\newblock {{A}pproximation by Superpositions of a Sigmoidal Function}.
\newblock Math. Control Signals Syst. 1989 2(4):303--314. 

\bibitem{F} 
Funahashi KI.
\newblock {{O}n the approximate realization of continuous mappings by neural networks}.
\newblock Neural Netw. 1989 2(3):183--192.

\bibitem{LS} 
Liang S, Srikant R.
\newblock {{W}hy deep neural networks for function approximation?}.
\newblock Proceedings of the 5th International Conference on Learning Representations (ICLR) 2017 
\newblock 2016 arXiv:1610.04161

\bibitem{HSW1} 
Hornik K, Stinchcombe M, White H.
\newblock {{M}ultilayer feedforward networks are universal approximators}.
\newblock Neural Netw. 1989 2(5):359--366

\bibitem{HSW2} 
Hornik K, Stinchcombe M, White H.
\newblock {{U}niversal approximation of an unknown mapping and its derivatives using multilayer feedforward networks}.
\newblock Neural Netw. 1990 3(5):551--560

\bibitem{P} 
Pinkus A.
\newblock {{A}pproximation theory of the MLP model in neural networks}.
\newblock Acta Numer. 1999 8:143--195.

\bibitem{ES} 
Eldan R, Shamir O.
\newblock {{T}he power of depth for feedforward neural networks}.
\newblock 29th Annual Conference on Learning Theory 2016 PMLR 49:907-940

\bibitem{S} 
Suzuki S.
\newblock {{C}onstructive function-approximation by three-layer artificial neural networks}.
\newblock Neural Netw. 1998 Aug;11(6):1049--1058.

\bibitem{M} 
Murata N.
\newblock {{A}n integral representation of functions using three-layered networks and their approximation bounds}.
\newblock Neural Netw. 1996 9(6):947--956

\bibitem{SM} 
Sonoda S, Murata N.
\newblock {{N}eural network with unbounded activation functions is universal approximator}.
\newblock Appl. Comput. Harmon. A. 2017 43(2):233--268. 

\bibitem{SM2} 
Sonoda S, Murata N.
\newblock {{T}ransport analysis of infinitely deep neural network}.
\newblock J. Mach. Learn. Res. 2019 20(2):1--52. 

\end{thebibliography}
\end{document}